\documentclass[twoside,11pt]{article}

\usepackage{xcolor}
\usepackage[ruled,vlined]{algorithm2e} 
\usepackage{amsmath,multirow} 
\usepackage{enumerate,array}
\usepackage{amssymb}
\usepackage[euler]{textgreek} 
\usepackage{amsmath} 
\usepackage{booktabs} 
\usepackage{adjustbox,lipsum} 
\usepackage{graphicx}

\SetKwInput{KwInput}{Input}                
\SetKwInput{KwOutput}{Output}              

\usepackage{jmlr2e}

\firstpageno{1}

\begin{document}

\title{Comparing effectiveness of regularization methods on text classification: Simple and complex model in data shortage situation }

\author{\name Jongga Lee \ \email closingprice@naver.com \\
       \addr Department of Applied Statistics\\
       Chung-Ang University \\
       Seoul, Korea
       \AND
       \name Jaeseung Yim \email yimyh1231@gmail.com \\
   	   \addr Department of Applied Statistic\\
   	   Chung-Ang University \\
   	   Seoul, Korea
   	   \AND
   	   \name Seo Hui Park \email hee30970@naver.com \\
   	   \addr Department of Applied Statistic\\
   	   Chung-Ang University \\
   	   Seoul, Korea
   	   \AND
   	   \name Changwon Lim \ \email clim@cau.ac.kr \\
   	   \addr Department of Applied Statistics\\
   	   Chung-Ang University \\
   	   Seoul, Korea
}

\editor{}

\maketitle

\begin{abstract}
Text classification is a task of assigning a document into one of predefined class. However it is expensive to acquire enough number of labelled documents, or to label them. In this paper, we study the regularization methods' effect on various classification models, when only few labelled data is available. We compare simple word embedding-based model which is simple but effective model and complex models (CNN and BiLSTM). In supervised learning, adversarial training can regularize the model further. When unlabelled dataset is available, we can regularize the model using semi-supervised learning methods such as Pi model and virtual adversarial training. We evaluates the regularization effects on 4 text classification datasets (AG news, DBpedia, Yahoo! Answers, Yelp Polarity), using only 0.1\%~0.5\% of its original labelled training documents. The simple model relatively performs well in fully supervised learning, but with help of adversarial training and semi-supervised learning, both simple and complex models can be regularzied, showing better results for complex models. Although the simple model is robust to overfit, complex model with well-designed priori belief can be also robust to overfit. 
\end{abstract}


\section{Introduction}

Text classification is a task which assigns a document to a predefined class. It is an important task in a knowledge system since substantial amount of textual data can be accumulated on the system. However, most text classification methods require a lot of labelled documents, which one may find it expensive to collect (Altınel et al., 2017). This is true especially when skilled workers, or experts, should be involved to label the literature. On the other hand, one can relatively acquire unlabeled documents easily.

In deep learning, it requires substantial amount of data for a model to be trained. For example, Yahoo! Answers dataset contains 1.4 million documents for 10 classes. This is due to a parametric aspect of deep learning models, which have tons of parameters to be trained (Vinyals et al., 2016). Therefore, in order to reduce the overfitting, deep learning has exploited various regularization methods, by tweaking layer such as adding dropout mask (Srivastava et al., 2014), or by adding regularization term to an objective function such as weight regularization. Adding regularization terms to an objective function, in bayesian perspective, provides a priori knowledge about the model (Bishop, 2006). For example, weight regularization encourages the model’s parameters not to be large. One can expect additional regularization effects with semi-supervised learning and adversarial training.

Semi-supervised learning takes advantage of unlabeled instances, from which the model calculates the unsupervised loss, such as entropy ( Grandvalet and Bengio, 2005). Laine and Aila (2017) use entropy loss and divergence loss, forcing robustness between two random perturbations, such as two different dropout masks. The method adds distribution smoothing term to the objective term, which correspond to the general knowledge that the distribution should be smooth over an input (Miyato et al., 2019). In adversarial training, the training makes use of an adversarial perturbation, which can be created by finding a perturbation that confuses the model’s classification loss most (Goodfellow et al., 2015). Compared to the common technique to add small random noises to inputs as a regularizer, the adversarial training suggests the direction toward which the model’s classification behavior is most sensitive. The adversarial training can be applied to a semi-supervised learning, which is called virtual adversarial training (Miyato et al., 2019). Unlike the original, virtual adversarial training finds the small noise which perturb the model’s output distribution. Initially applied to image classification task, this method was applied to text classification by taking word embeddings as an input (Miyato et al., 2016). 

A simple but intuitive alternative to tackle with data shortage is to use a simpler model, since a complex model cannot generalize well to small dataset and learns irrelevant noises. This is known as bias-variance dilemma (Luxburg et al., 2011). Shen et al. (2018) suggests a simple word embedding-based model, which only uses pooling operations over word embeddings. Despite of its simplicity, the model shows a comparable (or even slightly better) performance in natural language processing tasks, including text classification. In addition, the authors show that the model’s performance is robust to the number of training data compared to more complex models, thanks to its simplicity. Since the simple word embedding models does as well as complex models, thus by occam’s razor, it can be said that a simple model is the best choice among candidate models. 

However, in statistics, it is crucial to select the proper model (Cox, 2006). Neal et al. (2018) shows that in modern neural networks, both bias and variance can decrease as the number of parameters grows, conflicting bias-variance dilemma. In text classification, Convolutional neural network (CNN) and Recurrent neural network (RNN) are popular. CNN uses a convolution operation over n-consecutive words, creating n-gram features (Shen et al., 2017). RNN encodes a document seqeuntaily, with its internal memory (Lai et al., 2015), having some variants depending on the formulation of the internal memory. These models seem more appropriate approaches for text classification judging by how they encode sentences. In the view of model selection, we guess that the simple model suggested by Shen et al. (2018) outperforms complex models, because the complex models are ill-regularized. Regularization is a technique of increasing performance of a model by reducing overfittng. 

In this paper, we apply regularization methods to text classificaiton models, both simple and complex models, when labelled data is limited. Data shortage can be represented by 1) only few labelled instances 2) few labelled instances but enough unlabelled instances. Therefore, we evaluate the effect in supervised and semi-supervised settings. Specific methods are introduced in section 2. The rest of this paper is organized as follows. Section 2 describes the methods used in this paper: adversarial training, semi-supervised learning and deep text classification models. In Section 3, the experiments are described and the results are provided. In Section 4, we discuss the results. In Section 5, we draw conclusion of this paper and suggest topics for the future research.

\section{Methods}
In this section, we describe the text classification model in deep learning, adversarial training and semi-supervised training process used in this paper. A matrix is written in a bold capital character, a vector in a bold lowercase character and a scalar in normal lowercase character.

\subsection{Model for text classification}
In deep learning, the model processing text uses word embeddings as an input. Suppose, a document consists of T words, then its input is ,  is a word vector. The model is made of two parts; a composition function  and a classifier .  takes , producing a vector . Then the classifier outputs a distribution, , where  is the number of classes.
The popular choice for  is LSTM (Hochreiter and Schmidhuber, 1997), BiLSTM (Schuster and Paliwal, 1997) or CNN. All of these network contain a set of parameters, while the simple word embedding-based model (SWEM) suggested by shen et al. (2018) does not. Its composition function is as follow:
\begin{eqnarray}
\textbf{z}=f(\textbf{X})=
\begin{cases} 
	\frac{1}{T} \sum_{i=1}^{T} \textbf{x}_{i},& (average \ pooling) \\ 
	\max_{i=1}^{T} \textbf{X}_{ij}, & (max \ pooling)
\end{cases}
\end{eqnarray}
\\
Although SWEM has a variant version, for example hierarchical pooling or concatenation both, reader can notice that any variants of SWEM do not need any parameters to encode the input matrix  into a vector. 

In this paper, we use four compositional function; concatenated SWEM, CNN, BiLSTM and BiLSTM(MAX). BiLSTM(MAX) is a bidirectional LSTM, but unlike other LSTM, max pooling is applied to hidden states over all timesteps. Since contexts up to any timestep can directly contribute to encoded features, we expect better learning over conventional BiLSTM.

\subsection{Regularization methods}
In this subsection, we introduce the regularization methods we used in this paper.\\

\noindent \textbf{Adversarial training}\\
An adversarial example is an input added with small and imperceptible noise, which causes the model to make an incorrect classification (Goodfellow et al., 2015). Goodfellow et al. (2015) suggested the new method to find this adversarial noise and used it for training. The main part of adversarial training is to find a noise, which disturbs (or maximizes) the model’s classification loss. Without actually solving any optimization problems, the adversarial noise can be calculated as a gradient of classification loss with respect to input, thanks to a linear property of neural networks. Adding random noise to inputs also has similar effects, smoothing the distribution. However, unlike random noise addition, adversarial noise is a direction toward which the model’s behavior is vulnerable, therefore adversarial training shows better regularization results. See Algorithm~\ref{alg:the_alg1}  for a the trainining process for details.

\begin{center}
\begin{algorithm}[H]
\DontPrintSemicolon
  \textbf{Notation:} \\ 
  \quad  J($\boldsymbol\theta$,$\mathbf{x}$,$y$) : cross entropy loss for a model paratemerzied with $\boldsymbol\theta$ \\
  \KwInput{\\ \ \ \ \ $\mathbf{x}$ is an input \\ \ \ \ \ $y$ is a corresponding label}
  \textbf{Algorithm:} \\
  \textbf{1.} Get a gradient of classification loss J with respect to input \textbf{x} \\
  \quad  \textbf{\texteta} = $\nabla_{\textbf{x}}$ J($\boldsymbol\theta$,$\mathbf{x}$,$y$) \\
  \textbf{2.} Normalize the \textbf{\texteta} \ and set its size to \textepsilon \\
  \quad \textbf{\texteta} = \textepsilon * $L_{2}$(\textbf{\texteta}) \tcp*{$L_{2}$ is a L2 normalization}
  \textbf{3.} Set the final objective function as weighted sum of classification loss between original and its adversarial example \\
 \quad  J($\boldsymbol\theta$,$\mathbf{x}$,$y$) = \textalpha *  J($\mathbf{\theta}$,$\mathbf{x}$,$y$) + (1-\textalpha)* J($\boldsymbol\theta$,$\mathbf{x}$+$\boldsymbol{\eta}$,$y$)
\caption{Adversarial training}
\label{alg:the_alg1}
\end{algorithm}
\end{center}

\noindent \textbf{Pi model}\\
Laine and Aila (2016) suggested a supervised learning method, called Pi model. This method adds a divergence term between two randomly perturbation, such as adding different random noises to an input, applying different dropout masks or different cropping if inputs are images. Mean squared error (MSE) was used as a divergence term. See Algorithm~\ref{alg:the_alg2} for the specific process. Note that it does not need any true label information to calculate a divergence term, but only its output distribution from the model is required. Divergence term encourages the model to be robust to random perturbations, making its distribution smooth over sample data points and their vicinity. Pi model has an intuitive concept, which is similar to ladder network (Rasmus et al., 2015) but simple and outperforms the ladder network in semi-supervised settings. Instead of a denoising task in the ladder network, in Pi  model, the divergence of model’s output distributions is a concern. Due to its simplicity, Pi model is applicable to text classification, since the concept of autoencoder is infeasible for text data.
In text classification, we mimic the perturbation by replacing words with “unknown” tokens, or switching the word order in a document. Note that for SWEM with max pooling, these perturbations do not have any effects. Not only for SWEM, the impacts of these text-specific perturbations were trivial.

\begin{center}
\begin{algorithm}[h]
\DontPrintSemicolon
  \textbf{Notation:} \\ 
  \quad  H($\mathbf{\theta}$,$\mathbf{x}$) : entropy loss functiom a model paratemerzied with $\mathbf{\theta}$ \\
  \quad  MSE($\mathbf{p_{1}}$,$\mathbf{p_{2}}$) : Mean squared error function between two distribution \\ \quad \qquad \qquad \quad \ \ \ \ \ $\mathbf{p_{1}}$,$\mathbf{p_{2}}$ \\
  \quad  p($\boldsymbol\theta$,$\mathbf{x}$) : a m model's output distribution ofr input $\mathbf{x}$, given its parameter $\boldsymbol\theta$ \\
  \KwInput{\\ \ \ \ \ $\mathbf{x_{l}}$ is an input with label $y$ \\ \ \ \ \ $\mathbf{x_{ul}}$ is an input without a label}
  \textbf{Algorithm:} \\
  \textbf{1.} Cross entropy loss for  $\mathbf{x_{l}}$ and entropy loss for $\mathbf{x_{ul}}$ \\
  \quad  L = J($\mathbf{\theta}$,$\mathbf{x_{l}}$,$y$) +  H($\mathbf{\theta}$,$\mathbf{x_{ul}}$) \\
  \textbf{2.} Get two randomly perturbed inputs from  $\mathbf{x_{ul}}$ and each output distributions \\
  \quad  $\mathbf{p^1}$=p($\boldsymbol\theta$,$\mathbf{x_{ul}^{1}}$) and $\mathbf{p^2}$=p($\boldsymbol\theta$,$\mathbf{x_{ul}^{2}}$) \\
  \textbf{3.} Get the final objective function as sum of all terms \\
  \quad  L=L+MSE($\mathbf{p^1}$,$\mathbf{p^2}$)
\caption{Pi model}
\label{alg:the_alg2}
\end{algorithm}
\end{center}
\raggedbottom

\noindent \textbf{Virtual adversarial training}\\
There have been attempts to use distribution smoothing techniques for semi-supervised learning, based on the belief that a classification model should assign same labels to close data points, including Pi model in Section 2.3. Motivated by Goodfellow et al. (2015), Miyato et al. (2019) suggested the virtual adversarial training (VAT), new method of adversarial training which can make use of unlabelled examples. The trick is to use Kullback-Leibler divergence instead of classification loss. See Algorithm~\ref{alg:the_alg3} and Algorithm~\ref{alg:the_alg4} for the specific training process. In this case, the training encourages the output distribution to be robust against the local noise, since divergence term is used, not cross entropy. The method can take advantages of unlabelled examples, applicable to a semi-supervised task.
The authors eluded the high computation cost for virtual adversarial noise, with fast approximation methods. Using a taylor expansion and power method, it can calculate the adversarial noise without a real optimization problem, as in Goodfellow et al. (2015). 
\raggedbottom
\begin{center}
\begin{algorithm}[H]
\DontPrintSemicolon
  \textbf{Notation:} \\ 
  \quad  KLD($\mathbf{p^1}$,$\mathbf{p^2}$) : Kullback-Leibler divergence between two distributions \\ \qquad \qquad \qquad \quad$\mathbf{p^1}$,$\mathbf{p^2}$ \\
  \quad  $\mathbf{r}$ is a (initial) gaussian random noise of same shpae with $\mathbf{x}$ \\
  \KwInput{\\ \ \ \ \ $\mathbf{x_{l}}$ is an input with label $y$ \\ \ \ \ \ $\mathbf{x_{ul}}$ is an input without a label}
  \textbf{Algorithm:} \\
  \textbf{1.} Cross entropy loss for  $\mathbf{x_{l}}$ and entropy loss for $\mathbf{x_{ul}}$ \\
  \quad  L = J($\mathbf{\theta}$,$\mathbf{x_{l}}$,$y$) +  H($\mathbf{\theta}$,$\mathbf{x_{ul}}$) \\
  \textbf{2.} Generate an adversarial perturbation \\
  \quad  $\mathbf{r_{vadv}}$=genVadv($\mathbf{r}$,p($\boldsymbol\theta$,$\mathbf{x_{ul}}$)) \tcp*{Feedforward here to be  masked}
  \textbf{3.} Get the final objective function by adding divergence term using the adversarial perturbation in 2. \\
  \quad  L=L+KLD(p($\boldsymbol\theta$,$\mathbf{x_{ul}}$),p($\boldsymbol\theta$,$\mathbf{x_{ul}}$+$\mathbf{r_{vadv}}$))
\caption{Virtual adversarial training}
\label{alg:the_alg3}
\end{algorithm}
\end{center}

\begin{center}
\begin{algorithm}[H]
\DontPrintSemicolon
  Generate virtual adversarial perturbation : genVadv($\mathbf{r}$,p($\boldsymbol\theta$,$\mathbf{x_{ul}}$)) \\ 
  \textbf{Notation:} \\ 
  \quad  KLD($\mathbf{p^1}$,$\mathbf{p^2}$) : Kullback-Leibler divergence between two distributions $\mathbf{p^1}$,$\mathbf{p^2}$ \\
  \quad  $\mathbf{r}$ is a (initial) gaussian random noise of same shpae with $\mathbf{x}$ \\
  \KwInput{\\ \ \ \ \ $\mathbf{x_{l}}$ is an input with label $y$ \\ \ \ \ \ $\mathbf{x_{ul}}$ is an input without a label}
  \textbf{Algorithm:} \\
  \textbf{1.} Normalize the initial random noise $\mathbf{r}$ \\
  \quad $\mathbf{r}$ = $L_{2}$($\mathbf{r}$) \\
  \textbf{2.} Compute a distribution for finite differentiation  \\
	
  \quad  $\mathbf{r_{vadv}}$=genVadv($\mathbf{r}$,p($\boldsymbol\theta$,$\mathbf{x_{ul}}$)) \tcp*{Feedforward here to be  masked}
  \textbf{3.} Get the final objective function by adding divergence term using the adversarial perturbation in 2. \\
  \quad  L=L+KLD(p($\boldsymbol\theta$,$\mathbf{x_{ul}}$),p($\boldsymbol\theta$,$\mathbf{x_{ul}}$+$\mathbf{r_{vadv}}$))
\caption{genVadv Module}
\label{alg:the_alg4}
\end{algorithm}
\end{center}
\section{Experiment and results}

\subsection{Datasets}

\begin{table}[h]
\centering
\begin{tabular}{|l|l|l|l|l|}
\hline
Dataset                                                 & \# Class & Avg length & \# Training & \# Test \\ \hline
AG news                                                 & 4        & 57         & 600         & 9600    \\ \hline
DBpedia                                                 & 14       & 43         & 1400        & 70000   \\ \hline
Yahoo! Answer & 10       & 104        & 1400        & 60000   \\ \hline
Yelp Polarity & 2        & 139        & 600         & 38000   \\ \hline
\end{tabular}
\caption{Dataset description}
\label{tab:table1}
\end{table}

We evaluate the effectiveness of regularization methods on 4 different text classification datasets, AG news, DBpedia, Yahoo! Answers and Yelp Polarity, all of which are used in Shen et al. (2018), originally constructed by Zhang et al. (2015). The length of the datasets, number of classes and tasks are diverse. Dataset information is listed in Table~\ref{tab:table1}. All the datasets are publicly available on the internet. For Yelp Polarity, we predict a binary label (positive or negative) regarding one review about a restaurant. AG news is a topic classification dataset constructed by choosing 4 largest classes from the original AG corpus, by Xiang Zhang (xiang.zhang@nyu.edu). DBpedia is extracted from Wikipedia by crowd-sourcing and is categorized into 14 non-overlapping ontology classes, including Company, Athlete, Natural Place, etc. Yahoo! Answer is a topic classification dataset of a set of question and its best answer, with 10 classes, such as Health, Sports and Politics \& Government.
In order to implement the data limited situations, we randomly sampled 0.5\% to 0.1\% of original dataset for training labelled instances, evenly from each categories. The unlabelled data could be abundant or minimal, so we used 20/10/5/2 times of the labelled data. Validation datasets are randomly selected as a half size of the test datasets, having even frequencies among categories. For example, the number of the labelled of AG news is reduced to 600 (150 per classes), 4800 validation dataset, and the number of unlabelled instances are set to  12,000/6,000/3,000/1,200 according to each scenario.

\subsection{Experiment setup}

\begin{table}[!]
\centering
\begin{tabular}{|c|c|}
\hline
Model       & Configuration                                                                              \\ \hline
BiLSTM(MAX) & Hidden\_state : 256                                                                        \\ \hline
BiLSTM      & Hidden\_state : 256                                                                        \\ \hline
CNN         & \begin{tabular}[c]{@{}c@{}}Num\_kernel : 300\\ Context\_size : 7\\ Stride : 2\end{tabular} \\ \hline
SWEM        & -                                                                                          \\ \hline
\end{tabular}
\caption{Model configuration}
\label{tab:table2}
\end{table}

We use GloVe (Pennington et al., 2014) for the word embedding initializer. The words not in GloVe’s vocabulary are initialized with a uniform distribution [-0.01, 0.01]. Dropout is applied after composition function and classifier, with same dropout rate among [0.2, 0.3, 0.5]. Configuration for each model is in Table~\ref{tab:table2}. Note that SWEM is a concatenated SWEM, since it has the best performance among SWEM variants. The classifier layer is MLP layer (relu) whose dimension is fixed to 300. Adam (Kingma and Ba, 2015) optimizer is used, and a set of candidates learning rate is [3e-4, 5e-4, 1e-3, 3e-3], without learning rate scheduler. 
All the methods are implemented with TensorFlow (Abadi et al., 2016), and NLTK (Bird et al., 2016) for text processing, such as a tokenizer. One TITAN X GPU 12GB memory is used for training.

\subsection{Experiment results}

\begin{table}
  \centering
    \begin{adjustbox}{max width=\textwidth}
    \begin{tabular}{|c|c|c|c|c|c|}
    \hline
    \multirow{2}{*}{DBpedia} & \multicolumn{2}{c|}{Supervised} & \multicolumn{3}{c|}{Semi\-supervised} \\ \cline{2-6} 
                             & sup          & at          & pi    & vat    & at+vat   \\ \hline
    BiLSTM(MAX)              & 94.02           & \textbf{95.97}         & \textbf{96.94/97.08/96.66/95.57}    & \textbf{97.26/96.93/96.41/95.50}     & \textbf{97.62/97.55/97.31/96.11}       \\ \hline
    BiLSTM                   & 90.31           & 93.63         & 94.93/94.70/93.63/87.61    & 95.01/94.52/93.58/88.76     & 95.01/94.52/93.58/88.76       \\ \hline
    CNN                      & 91.71           & 92.69         & 96.59/96.18/95.64/93.22    & 96.94/96.12/95.95/92.01     & 96.94/96.66/96.13/92.70       \\ \hline
    SWEM                     & \textbf{94.23}           & 94.38         & 95.98/95.68/95.39/94.58    & 96.10/95.39/95.02/94.11     & 96.15/95.76/95.47/95.34       \\ \hline
    \end{tabular}
    \end{adjustbox}
    \medskip

    \begin{adjustbox}{max width=\textwidth}
    \begin{tabular}{|c|c|c|c|c|c|}
    \hline
    \multirow{2}{*}{Yahoo! Answer} & \multicolumn{2}{c|}{Supervised} & \multicolumn{3}{c|}{Semi\-supervised} \\ \cline{2-6} 
                             & sup          & at          & pi    & vat    & at+vat   \\ \hline
    BiLSTM(MAX)              & 60.85           & \textbf{62.76}         & \textbf{65.43/64.85/65.04/63.32}    & \textbf{67.10/66.48/65.50/65.06}     & textbf{66.78/66.91/66.18/65.87}       \\ \hline
    BiLSTM                   & 54.58           & 58.14         & 61.54/61.71/60.87/57.65    & 63.05/63.40/62.37/59.64     & 65.79/65.93/65.52/62.39       \\ \hline
    CNN                      & 57.06           & 59.88         & 62.56/61.95/61.97/59.84    & 63.32/63.43/62.37/61.77     & 65.13/64.59/64.23/63.08       \\ \hline
    SWEM                     & \textbf{61.34}           & 61.34        & 63.99/63.56/63.41/62.60    & 62.24/63.81/63.37/63.13     & 62.79/62.20/60.83/61.36       \\ \hline
    \end{tabular}
    \end{adjustbox}
    \medskip

    \begin{adjustbox}{max width=\textwidth}
    \begin{tabular}{|c|c|c|c|c|c|}
    \hline
    \multirow{2}{*}{Yelp Polarity} & \multicolumn{2}{c|}{Supervised} & \multicolumn{3}{c|}{Semi\-supervised} \\ \cline{2-6} 
                             & sup          & at          & pi    & vat    & at+vat   \\ \hline
    BiLSTM(MAX)              & \textbf{87.32}           & \textbf{88.21}         & \textbf{86.94/88.46/86.98/88.26}     & \textbf{90.10/88.84/88.20/88.61}      & \textbf{89.85/89.16/89.11/88.55}        \\ \hline
    BiLSTM                   & 71.66           & 79.12         & 75.78/78.57/74.93/80.03    & 75.63/76.95/73.42/74.71     & 87.25/86.59/87.62/86.47       \\ \hline
    CNN                      & 81.34           & 84.07         & 83.44/81.99/81.41/80.47    & 77.26/82.69/57.47/58.40    & 87.93/85.57/85.93/85.40       \\ \hline
    SWEM                     & 81.34           & 83.00         & 83.09/77.25/81.27/81.47    & 77.26/83.76/87.39/87.20    & 87.57/87.23/87.33/86.98       \\ \hline
    \end{tabular}
    \end{adjustbox}
    \medskip

    \begin{adjustbox}{max width=\textwidth}
    \begin{tabular}{|c|c|c|c|c|c|}
    \hline
    \multirow{2}{*}{AG news} & \multicolumn{2}{c|}{Supervised} & \multicolumn{3}{c|}{Semi\-supervised} \\ \cline{2-6} 
                             & sup          & at          & pi    & vat    & at+vat   \\ \hline
    BiLSTM(MAX)              & 81.15           & 82.96         &84.68/85.39/84.11/\textbf{85.22}    & 85.38/\textbf{86.38/85.48}/84.92     & \textbf{87.27/86.98/86.09/86.09}       \\ \hline
    BiLSTM                   & 77.40           & 78.90         & 85.35/85.22/83.65/83.44    &  85.94/85.34/85.22/\textbf{85.28}     & 86.31/85.90/85.26/84.30      \\ \hline
    CNN                      & 78.90           & 81.02         &85.55/85.02/84.43/81.01    & \textbf{86.34}/85.61/85.02/82.84     & 85.89/85.94/85.46/82.50       \\ \hline
    SWEM                     & \textbf{82.61}           & \textbf{83.53}         & \textbf{86.02/86.38/85.68}/84.71    & 85.90/85.84/85.35/83.89     & 85.28/84.56/84.86/83.81       \\ \hline
    \end{tabular}
    \end{adjustbox}
  \caption{Experiments results on 4 datasets}
  \label{tab:table3}
\end{table}

Results are shown in Table~\ref{tab:table3}, listed in 4 tables, one for each dataset. The values are accuracy (\%). “Sup” denotes fully supervised learning, “AT” for adversarial training, “Pi” for pi model, “VAT” for virtual adversarial training and “AT+VAT” for adversarial training (labelled) and virtual adversarial training (unlabelled). In “semi-supervised”, results with different number of unlabelled instances are separated by a slash
Let us begin with DBpedia. In supervised learning, SWEM achieved higher accuracy compared to other models, especially CNN and BiLSTM. Even with adversarial training, SWEM is still better, while in semi-supervised settings, the performance gaps are much narrowed. CNN and BiLSTM gain much benefits from regularization. BiLSTM(MAX) shows comparable results to SWEM even in supervised learning (Sup) and achieved higher accuracy with adversarial training (AT). In addition, with help of semi-supervised learnings, BiLSTM(MAX) achieve the best performance, 97.62\% with adversarial training plus virtual adversarial training (AT+VAT). Next, we explore the effect of the number of unlabelled instances in semi-supervised learning. It seems that even with twice of the labelled instances, the performance gain over supervised learning (Sup and AT) can be observed. For example, for BiLSTM(MAX), 95.57\% for adversarial training (AT) but 96.11\% for adversarial training plus virtual adversarial training (AT+VAT). However, more unlabelled instances leads to slightly better results.
The similar pattern can be found in other datasets. SWEM achieved higher accuracy in supervised learning (Sup and AT), while with adversarial training and semi-supervised learning, BiLSTM(MAX) achieved the highest accuracy. In semi-supervised learning, the performance gain grows as the number of unlabelled instances increases in semi-supervised learning. However, in AG news the patterns are relatively unclear, and we check the variance of results in discussion section.

\section{Discussion}
\subsection{Distribution smoothing and consistent training}

\begin{table}
  \centering
  \begin{adjustbox}{max width=\textwidth}
  \begin{tabular}{|c|c|c|c|}
  \hline
  Model & Mean & Std & Max-Min \\ \hline
  BiLSTM(MAX) & 80.40/83.53/85.79/86.49/87.28 & 1.29/0.841/0.597/ 0.273/0.122 & 4.26/3.26/2.40/ 0.697/0.381 \\ \hline
  BiLSTM & 76.41/79.13/84.99/85.72/86.02 & 2.41/1.56/0.495/ 0.382/0.308 & 9.09/4.93/1.73/ 1.184/0.973 \\ \hline
  CNN & 77.75/80.05/85.41/85.99/86.33 & 1.52/1.14/0.218/ 0.279/0.218 & 5.052/4.157/ 0.7368/0.697 \\ \hline
  SWEM &  81.17/82.46/86.35/86.21/86.22 & 1.07/0.721/0.446/0.113/0.129 & 3.76/2.35/1.35/ 0.434/0.513 \\ \hline
  \end{tabular}
  \end{adjustbox}
  \caption{10 Repeated results in AG news}
  \label{tab:table4}
\end{table}

\begin{figure}[!tbp] 
  \centering
  \begin{minipage}[b]{0.4\textwidth}
    \includegraphics[width=\textwidth]{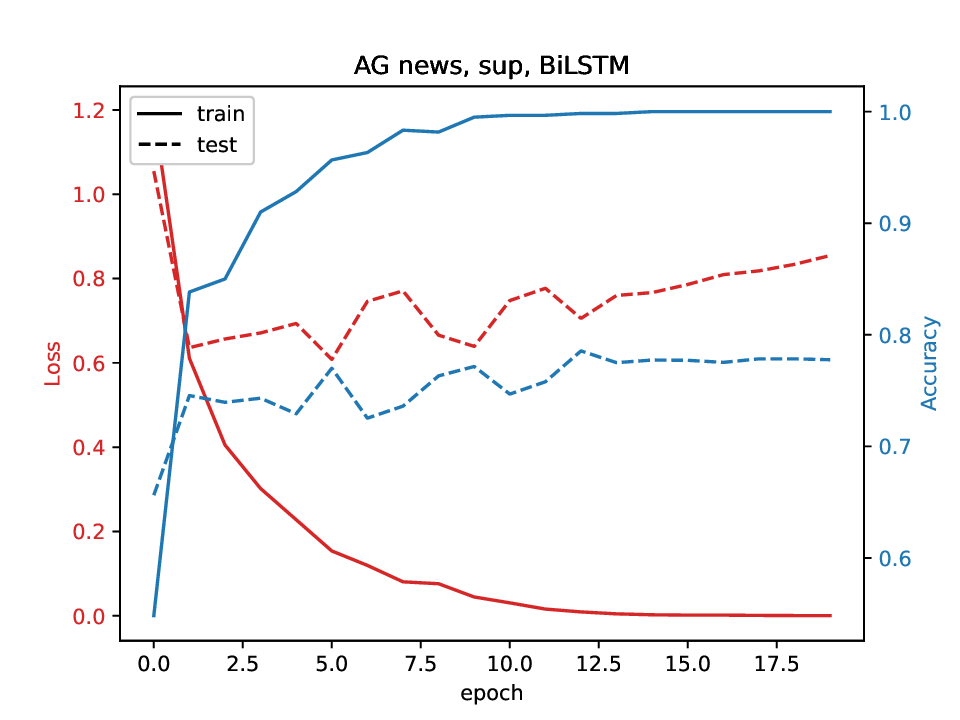}
  \end{minipage}
  \begin{minipage}[b]{0.4\textwidth}
    \includegraphics[width=\textwidth]{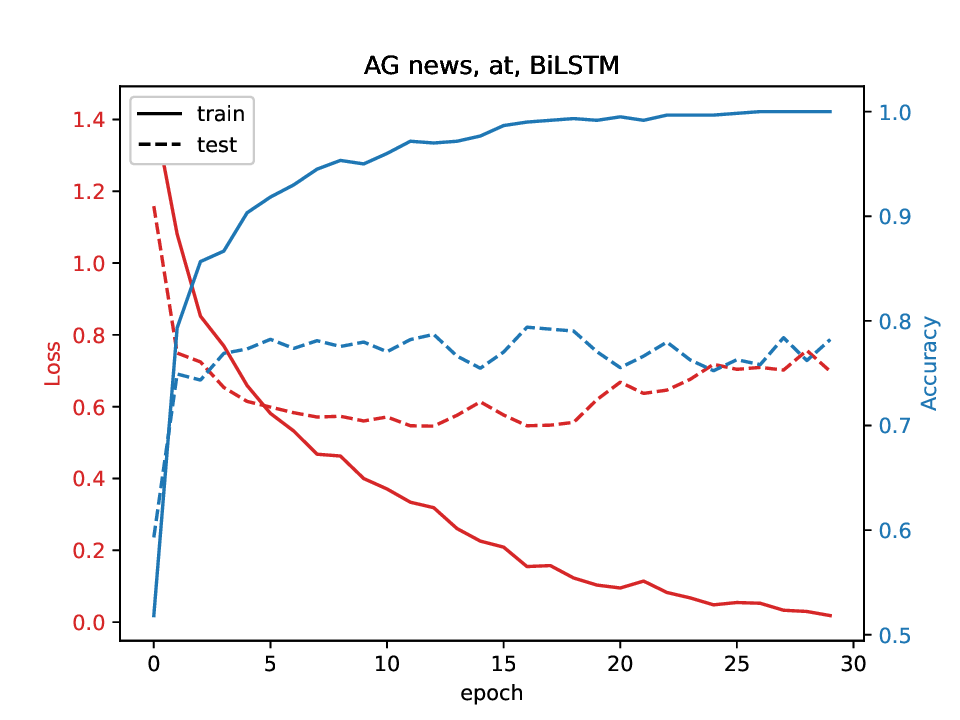}
  \end{minipage}
  \caption{Training procees with and without regularization}
  \label{fig:figure1}
\end{figure}

In our experiments, we found that only in AG news, the patterns are mixed, while in other datasets, the patterns look clear. We guess that the variance of results are relatively high, thus we repeat experiments 10 times. See Table~\ref{tab:table4}. Sup, AT, Pi, VAT and AT+VAT are separated by a slash, and semi-supervised learning are trained with 20 times unlabelled instances. Mean, standard deviation and gap between max and min are listed. With distribution smoothing, the trained model can achieve not only a better accuracy, but a lower variance.
Smoothing methods also lead to a stable training, as in Figure~\ref{fig:figure1}. In fully supervised learning, the validation accuracy reaches to its best value at only 3 epochs, indicating further training makes worse generalization of model. On the other hand, one trained with a distribution regularization shows a flattened stage through the training process. 

\subsection{LSTM with max pooling}
\begin{figure}[!tbp]
	\caption{Frequency of timesteps contributing to final feature}
	\label{fig:figure2}
	\begin{center}
		\includegraphics[scale=0.5]{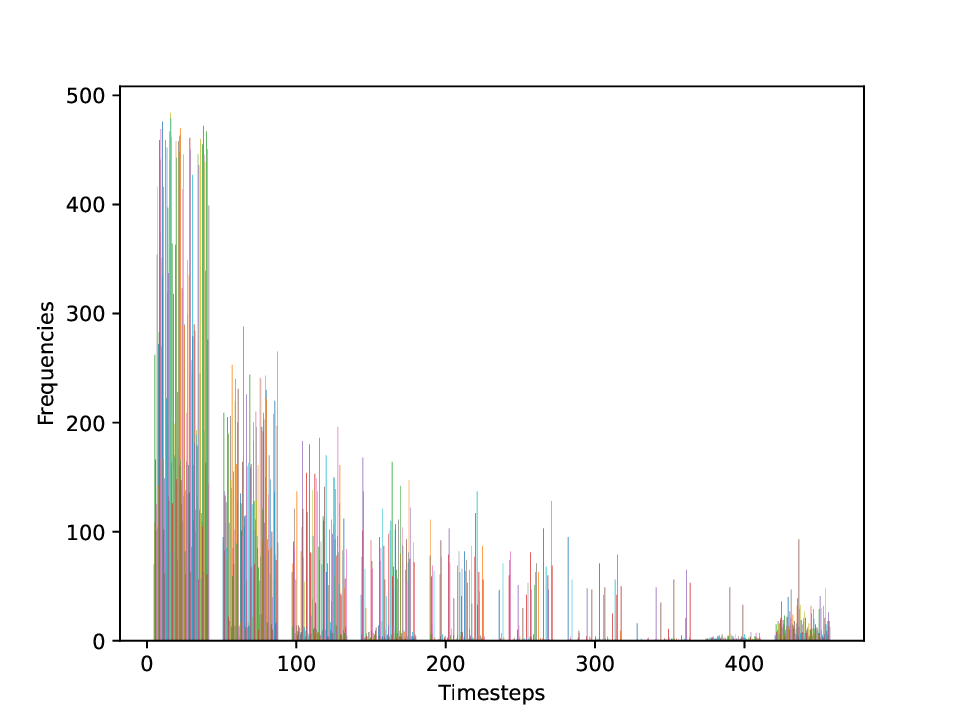}
		 
	\end{center}
\end{figure}

We conjectured that biLSTM(MAX) showed better performances than BiLSTM, because a hidden state at a certain timestep has a chance to contribute to the features of a document. We draw a histrogram, showing from which max values come from. See Figure \ref{fig:figure2}. It shows the frequency of timesteps contributing to final feature vector, for a single minibatch of size 128. Each colored line refers to each example in the minibatch(here, 128 different colors). We find that the final feature vector is constructed from a range of timesteps.

\section{Conclusion}
In this paper, we compare the regularization effects on a simple model and complex models for text classification when the labelled data is limited. Simple word embedding-based model is a simple model performing well in fully supervised learning, thanks to its simple model formulation. On the other hand, complex models can take much  leverage from regularization, especially the method using unlabelled instances. With the regularization using sensitive direction of the model’s behaviors, and with a bidirectional LSTM with pooling, the model performs decently even with a few labelled instances. 
Not only the improved performance, the regularization provides  consistent performances. Since regularization provides a distribution smoothing priori knowledge in bayesian view, it encourages the stability to the training process, compared to somewhat unstable training process in fully supervised training. This leads to consistent results for models. 
  Therefore, when available labelled dataset is small in size, from the experiment results in this paper, we conclude that model formulation is more important in text classification. Researchers can make use of both complex models and regularization methods, instead of using simple model. Furthermore, regularization could be applied to much complex model, such as Transformer (Vaswani et al., 2017).

\acks{This research was supported by Next-Generation Information Computing Development Program through the National Research Foundation of Korea (NRF) funded by the Ministry of Science, IC (2017M3C4A708328123). }

\bibliography{references}

\noindent Abadi, M., Barham, P., Chen, J., Chen, Z., Davis, A., Dean, J., … Kudlur, M. (2016). TensorFlow: A system for large-scale machine learning. \textit{Proceedings of the 12th USENIX Symposium on Operating Systems Design and Implementation}, 265-283. \\

\noindent Altınel, B., Can G. M., Diri, B. (2017). Instance labeling in semi-supervised learning with meaning values of words. \textit{Engineering Applications of Artificial Intelligence}, 62, 152-163, https://doi.org/ 10.1016/j.engappai.2017.04.003. \\

\noindent Bird, S. Klevin, E., Loper, E. (2009). \textit{Natural Language Processing with Python}. O'Reilly Media, Inc. \\

\noindent Bishop, C. M. (2006). \textit{Pattern Recognition and Machine Learning}. New York: Springer. \\

\noindent Cox, D. R. (2006). Principles of statistical inference. In \textit{Principles of Statistical Inference}. https://doi.org/10.1017/CBO9780511813559. \\

\noindent Goodfellow, I. J., Shlens, J., Szegedy, C. (2015). Explaining and harnessing adversarial examples. \textit{arXiv preprint} arXiv:1412.6572. \\

\noindent Grandvalet, Y., Bengio, Y. (2005). Semi-supervised learning by entropy minimization. \textit{Advances in Neural Information Processing Systems}, 529–536. \\

\noindent Hochreiter, S., Schmidhuber, J. (1997). Long Short-Term Memory. \textit{Neural Computation}, 9(8), 1735-1780. https://doi.org/10.1162/neco. 1997.9.8.1735. \\

\noindent Kingma, D. P., Ba, J. L. (2015). Adam: A method for stochastic optimization. \textit{3rd International Conference on Learning Representations}, ICLR 2015 - Conference Track Proceedings. \\

\noindent Lai, S., Xu, L., Liu, K., Zhao, J. (2015). Recurrent convolutional neural networks for text classification. \textit{Proceedings of the National Conference on Artificial Intelligence}, 2267–2273. \\

\noindent Laine, S., Aila, T. (2017). Temporal ensembling for semi-supervised learning. \textit{5th International Conference on Learning Representations}, ICLR 2017 - Conference Track Proceedings. \\

\noindent Miyato, T., Dai, A. M., Goodfellow, I. (2016). Adversarial training methods for semi-supervised text classification. \textit{arXiv preprint arXiv:1605.07725}.  \\

\noindent Miyato, T., Maeda, S. I., Koyama, M., Ishii, S. (2019). Virtual Adversarial Training: A Regularization Method for Supervised and Semi-Supervised Learning. \textit{IEEE Transactions on Pattern Analysis and Machine Intelligence}, 41(8) 1979-1993. https://doi.org/10.1109/ TPAMI.2018.2858821.  \\

\noindent Neal, B., Mittal, S., Baratin, A., Tantia, V., Scicluna, M., Lacoste-Julien, S., Mitliagkas, I. (2018). A modern take on the bias-variance tradeoff in neural networks. \textit{arXiv preprint} arXiv:1810.08591. \\

\noindent Pennington, J., Socher, R., Manning, C. D. (2014). GloVe: Global vectors for word representation. \textit{EMNLP 2014 - 2014 Conference on Empirical Methods in Natural Language Processing, Proceedings of the Conference}, 1532–1543. https://doi.org/10.3115/v1/d14-1162. \\

\noindent Rasmus, A., Valpola, H., Honkala, M., Berglund, M., Raiko, T. (2015). Semi-supervised learning with Ladder networks. \textit{Advances in Neural Information Processing Systems}, 3546–3554. \\

\noindent Schuster, M., Paliwal, K. K. (1997). Bidirectional recurrent neural networks. \textit{IEEE Transactions on Signal Processing}, 45(11), 2673-2681. https://doi.org/10.1109/78.650093. \\

\noindent Shen, D., Min, M. R., Li, Y., Carin, L. (2017). Learning context-sensitive convolutional filters for text processing. \textit{arXiv preprint} arXiv:1709.08294. \\

\noindent Shen, D., Wang, G., Wang, W., Min, M. R., Su, Q., Zhang, Y., Li, C., Henao, R., Carin, L. (2018). Baseline needs more love: On simple word-embedding-based models and associated pooling mechanisms. \textit{ACL 2018 - 56th Annual Meeting of the Association for Computational Linguistics, Proceedings of the Conference (Long Papers)}, 440–450. https://doi.org/10.18653/v1/p18-1041. \\

\noindent Srivastava, N., Hinton, G., Krizhevsky, A., Sutskever, I., Salakhutdinov, R. (2014). Dropout: A simple way to prevent neural networks from overfitting. \textit{Journal of Machine Learning Research}, 15(56), 1929-1958. \\

\noindent Vaswani, A., Shazeer, N., Parmar, N., Uszkoreit, J., Jones, L., Gomez, A. N., Kaiser, Ł., Polosukhin, I. (2017). Attention is all you need. \textit{Advances in Neural Information Processing Systems}, 6000–6010. \\

\noindent Vinyals, O., Blundell, C., Lillicrap, T., Kavukcuoglu, K., Wierstra, D. (2016). Matching networks for one shot learning. \textit{Advances in Neural Information Processing Systems}, 3637–3645. \\

\noindent von Luxburg, U. Schölkopf, B. (2011). Statistical learning theory: Models, concepts, and results. \textit{Handbook of the History of Logic}, 10, 651-706. https://doi.org/10.1016/B978-0-444-52936-7.50016-1. \\

\noindent Zhang, X., Zhao, J., Lecun, Y. (2015). Character-level convolutional networks for text classification. \textit{Advances in Neural Information Processing Systems}, 649–657.

\end{document}